\documentclass{l4dc2020} 

\title[Practical Reinforcement Learning For MPC]{Practical Reinforcement Learning For MPC:\\ Learning from sparse objectives in under an hour on a real robot}
\usepackage{times}
\usepackage{subcaption}

\author{%
 \Name{Napat Karnchanachari}$\,^1$ $^2$
 \thanks{Both authors contributed equally. $^1$ Sevensense Robotics A.G. $^2$ Robotic Systems Lab, ETH Zurich. $^3$ NVIDIA}
 \Email{patk@ethz.ch}
 \AND
 \Name{Miguel I. Valls}$\,^1$
 \footnotemark[1]
 \Email{miguel.delaiglesiavalls@sevensense.ch}
 \AND
 \Name{David Hoeller    }$\,^2$ $^3$
 \Email{dhoeller@ethz.ch}
 \AND
 \Name{Marco Hutter}$\,^2$
 \Email{mahutter@ethz.ch}
}
\begin{document}

\maketitle
\vspace{-0.4cm}

\begin{abstract}
Model Predictive Control (MPC) is a powerful control technique that handles constraints, takes the system's dynamics into account, and optimizes for a given cost function. In practice, however, it often requires an expert to craft and tune this cost function and find trade-offs between different state penalties to satisfy simple high level objectives.
In this paper, we use Reinforcement Learning and in particular value learning to approximate the value function given only high level objectives, which can be sparse and binary. Building upon previous works, we present improvements that allowed us to successfully deploy the method on a real world unmanned ground vehicle. Our experiments show that our method can learn the cost function from scratch and without human intervention, while reaching a performance level similar to that of an expert-tuned MPC. We perform a quantitative comparison of these methods with standard MPC approaches both in simulation and on the real robot. \\
A demonstration of our method can be seen in the video: \url{https://youtu.be/PJB8XdXBP_M}
\end{abstract}

\begin{keywords}
  Reinforcement Learning, Model Predictive Control, Autonomous Robots
\end{keywords}

\section{Introduction}
Model Predictive Control (MPC) is a trajectory optimization technique that has gained immense popularity over the last decades due to its ability to tackle inherently hard control problems~(\cite{Lee2011}). The theory is well understood and it is proven to be stable and optimal for a large variety of systems~(\cite{Lee2011}). MPC has been widely adopted due to algorithmic and technological advances. It can run in real time on a robot's on-board computing unit, allowing for applications such as autonomous racing (\cite{kabzan2019amz}), aggressive flight maneuvers with drones (\cite{mueller2013Drone}), and legged locomotion (\cite{neunert2018quadruped}).

In practice, however, it is well known that the cost functions for MPC have to be tuned. Experts craft specific costs that are a proxy for the original high level objectives, but also use costs that help the optimization converge, and avoid exploiting unmodelled or uncertain system dynamics. We refer to the latter as regularization cost terms. Finding a trade-off among regularization and proxy costs can be extremely difficult and time consuming (\cite{garriga2010model}). 

On the other side of the spectrum, Reinforcement Learning (RL) has shown to be a powerful tool, not only capable of handling binary and sparse rewards, but also overcoming the credit assignment problem when dealing with long horizons (\cite{silver2016mastering, lillicrap2015continuous, schulman2017proximal}). 

Recent approaches such as Plan Online, Learn Offline (POLO, \cite{lowrey2018plan}) or Deep Value Model Predictive Control (DMPC, \cite{farshidian2019deep}) attempt to combine best of both worlds by employing trajectory optimization with value function estimation. In this paper, we extend these works and learn to solve tasks defined by simple and easily interpretable high level objectives on a real unmanned ground vehicle (UGV). This is reputably challenging for most MPC algorithms because such objectives are represented by sparse and binary rewards. With this method, we can exploit our knowledge of the system dynamics and also endow the optimizer with a representation of the value landscape to solve the task in a sample efficient manner.
The main contributions of this paper are:
\begin{itemize}
    \item Presenting a practical extension of deep value function learning that outperforms a baseline MPC and is comparable to an expert-tuned MPC, trained from scratch on the real physical system in under 30 min with on-board CPU only.
    \item Showing that such learning based methods can be trained from high level binary and sparse rewards only, in under an hour, outperforming hand-tuned dense rewards learning.
    \item A comparison of two learning based techniques with standard MPC algorithms on a dynamical system with an uncertain model, in simulation and on a real system for trajectory tracking.
\end{itemize}

In the remainder of this paper, we introduce the background in Section~\ref{sec:background}, describe the method in Section~\ref{sec:method} and then proceed with the experiments in Section~\ref{sec:experiments}. We conclude with a review on related work in Section~\ref{sec:related_work} and the conclusion in Section~\ref{sec:conclusion}.
\begin{figure}[t!]
    \centering
    \includegraphics[width=\textwidth]{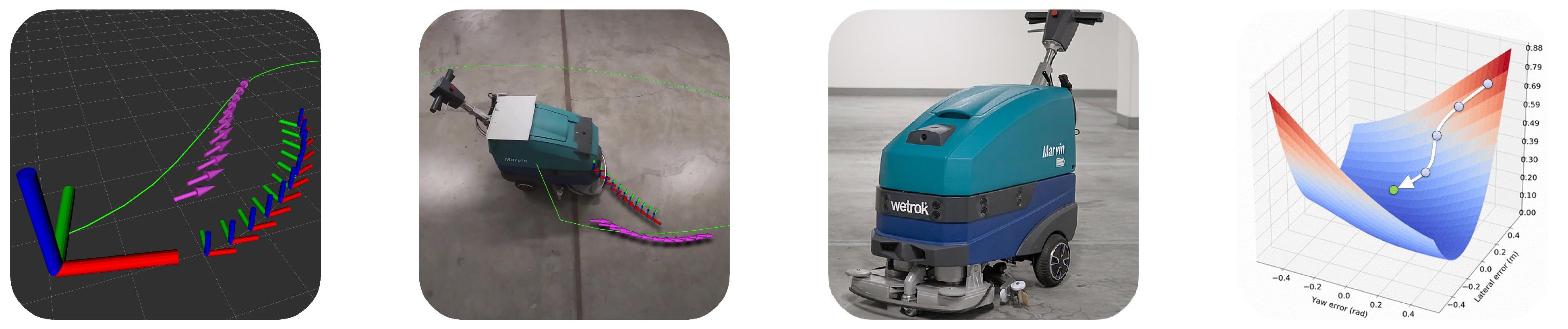}
    \caption{From left to right: simulation of the UGV, the real life UGV platform with visualization overlayed, and lastly an example of a 3D representation of the learned value function.}\label{fig:teaser}
\vspace{-0.5cm}
\end{figure}
\section{Background}\label{sec:background}
\subsection{Notation and Definitions}
In this problem setting, we consider an agent within an environment $E$ whose task is to maximize the discounted expected sum of rewards collected from the current time onwards. This is modelled by a Markov Decision Process (MDP) described by the state $\boldsymbol{s}_{t}  \in \mathbb{R}^{n_{s}}$ and dynamics $d\boldsymbol{s} = f(\boldsymbol{s}_{t}, \boldsymbol{a}_{t})$. At each time step, the agent observes a state $\boldsymbol{s}_{t}$ and takes an action  $\boldsymbol{a}_{t} \in \mathbb{R}^{n_{a}}$ according to the policy $\pi(\boldsymbol{s}_{t})$. The environment returns the corresponding evolved observations $\boldsymbol{s}_{t+1}$
, as well as a reward $r_t \in \mathbb{R}$. This forms a transition tuple $\langle\boldsymbol{s}_{t}, \boldsymbol{a}_{t}, r_{t}, \boldsymbol{s}_{t+1}\rangle$ for one time step. All transitions are stored in a replay buffer $\mathcal{B}$. The return is defined by the discounted future rewards $R = \sum^{\infty}_{t=0} \gamma^{t} r_{t}(\boldsymbol{s}_{t})$, where the discount factor is $\gamma \in [0,1)$. The value function $V(s)$ describes the expected return of being in state $s$, i.e. $V(s_0) = \mathbb{E}_{\pi}[R|\, s = \boldsymbol{s}_{0}]$. $V(s)$ is assumed to be time independent. The term cost is used as negative value, rendering minimizing cost an maximising value equivalent statements.

\subsection{Model Predictive Control}\label{sec:mpc}
MPC is a receding horizon control technique that maximizes a value function with respect to a sequence of control actions $\boldsymbol{a}_{0}\dots \boldsymbol{a}_{N-1}$ along a horizon of length $N$. The problem is constrained under state dynamics $\hat{f}(\boldsymbol{x}_{t}, \boldsymbol{a}_{t})$, and state and input constraints. Note that here, $\boldsymbol{x}\in \boldsymbol{S}_k \subseteq \text{R}^{n_s}$ and $\hat{f}(\boldsymbol{x}_{t}, \boldsymbol{a}_{t})$ denote the state and state dynamics of the actor respectively, which are distinguished from the true state and state dynamics of the environment. This results in the optimization
\begin{align}
 \pi_{MPC}(\boldsymbol{x}_0) &= \underset{\boldsymbol{a}_{0},\dots, \boldsymbol{a}_{N-1}}{\arg\max}  \, \gamma^{N}g_{N}(\boldsymbol{x}_{N}) + \sum^{N-1}_{k=0} \gamma^{k} g(\boldsymbol{x}_{k}, \boldsymbol{a}_{k})\label{eq:mpc_cost}\\[1em] 
 \text{s.t.} \quad\quad & \partial_{t}x_k = \hat{f}(\boldsymbol{x}_k, \boldsymbol{a}_k),\ \boldsymbol{a}_k \in \boldsymbol{A}_k, \,\boldsymbol{x}_k \in \boldsymbol{S}_k,\ \text{for } k = 0, \ldots, N, \nonumber
\end{align}
where $\boldsymbol{x}_0\in\text{R}^{n_s}$ denotes the current state feedback and $\boldsymbol{a}_k \in \boldsymbol{A}_k \subseteq \text{R}^{n_a}$ the control action, of which only the first is applied. This is then repeated for every control cycle. Notice that this MPC strategy is approximating the initial problem as a finite horizon optimal control problem, which depending on the horizon choice, is not guaranteed to be stable (\cite{Lee2011}). However, as will be seen in Section~\ref{sec:actor}, defining the terminal ($g_{N}(\cdot)$) and stage ($g(\cdot)$) costs in terms of the value function allows us to alleviate this limitation.

\subsection{Value Function learning}
Value function learning is commonly employed in reinforcement learning problems (\cite{sutton2016reinforcement}). In most cases, the value is represented by a state value function it $V_{\theta}(\boldsymbol{s}_{t})$ or an state action value function $Q_{\theta}(\boldsymbol{s}_{t}, \boldsymbol{a}_{t})$, parameterized by parameter vector $\theta$. This function is optimized minimizing the mean squared error loss with respect to a learning target, $y_{t}$ :
\begin{equation}
\label{eq:value_function}
    L = \mathbb{E}_{\pi_{\theta}}[(y_{t}-V_{\theta}(\boldsymbol{s}_{t}))^{2}].
\end{equation}

\section{Method}\label{sec:method}
The presented method is based off the actor-critic framework. The critic captures the global value function represented by a neural network. The actor is represented by a non-linear model predictive controller. The focus lays on the contributions that make it possible to run on a real physical system. The learning algorithm can be found in Appendix \ref{sec:training_algorithm}.

\subsection{Critic}
The key components to training the critic that are listed next. Note, this does introduce a new set of hyper parameters but were observed to have marginal effect compared to the value function.
\paragraph{Gradient regularization}
In the value function loss (Equation~\ref{eq:value_function}), we also regularize the Jacobian of the network. Since the Jacobian, $\boldsymbol{J} = \partial_{\boldsymbol{s}} V_{\theta}(\boldsymbol{s})$, and the Hessian of the network are required on the actor side (See Section~\ref{sec:qp_approx}), adding such a term will favor a smoother class of functions, which are easier to optimize for QP solvers.
Note that the Hessian is often approximated as $\boldsymbol{J}^T \boldsymbol{J}$ in the Gauss-Newton Hessian approximation (\cite{bjorck1996numerical}). Therefore, regularizing the Jacobian norm indirectly reduces the Hessian norm, and also aids numerical stability while training. Without this regularization, we found that running MPC with the learned value while training often did not converge. This resulted in unstable behavior, making it difficult to run on a real system.
\paragraph{Experience replay and data augmentation}
Experience replay is advantageous in two ways: it increases sample efficiency, and it stabilizes neural network training. This is realized in the form of replay buffer $\mathcal{B}$, see (\cite{silver2014deterministic}). Moreover, we augment our data profiting from the symmetries of the system (e.g. $V_{\theta}(\boldsymbol{x}) = V_{\theta}(h(\boldsymbol{x}))$). This not only implies that we can augment our data by the number of symmetries in the system but also that the agent will behave similarly well for symmetric states despite not having visited some of them.

\paragraph{n-step target}   
An n-step target is employed, it is the bootstrapped sum of discounted rewards over n consecutive steps (\cite{sutton2016reinforcement}):
\begin{equation}
    R^{(n)}(\boldsymbol{s}_{t}) = \sum^{n-1}_{i=0} \gamma^{i}r(\boldsymbol{s}_{t+i}) + \gamma^{n}V_{\theta}(\boldsymbol{s}_{t+n}) \label{eq:n_step_target}.
\end{equation}
This choice is due to its ability to balance bias and variance which affects Monte-Carlo and TD(0) returns respectively and in practice, it accelerates convergence. 

\paragraph{Target network}
We maintain a target network $V_{\theta'}(s)$ along side the critic network $V_{\theta}(s)$. Effectively, the Polyak-averaged version of the critic's estimated value function is used for bootstrapping. As shown by \cite{lillicrap2015continuous}, this trick greatly improves the actor-critic learning interaction.  
\subsection{Actor}\label{sec:actor}
RL and MPC can be combined in several ways further explained in Section~\ref{sec:related_work}. In this section we focus on two methods that employ value function learning.
\paragraph{Terminal Deep Value MPC}
The first and most intuitive combination (presented in \cite{lowrey2018plan}) uses the value function as terminal cost for the MPC. They show that bootstrapping the trajectory optimizer with the value function enables it to find global optimal solutions. Indeed, the value provides the missing information about the expected return from the end of the optimization horizon onwards. In this formulation, Equation~\ref{eq:mpc_cost} takes the following form:
\begin{align}
\pi_{TDMPC}(\boldsymbol{x}_0) &= \underset{\boldsymbol{a}_{0}\dots \boldsymbol{a}_{N-1}}{\arg\max}  \gamma^{N} V_{\theta}(\boldsymbol{x}_{N}) + \sum^{N-1}_{k=0} \gamma^{k} g(\boldsymbol{x}_{k}, \boldsymbol{a}_{k}) \label{eq:terminal_deep_mpc_cost}.
\end{align}
\cite{lowrey2018plan} use MPPI~(\cite{mppi2017}) to solve the MPC problem. Here, we use a Sequential Quadratic Programming (SQP) formulation that offers several key advantages, which are detailed later in this section. In the experiments, we refer to this variant as TDMPC.

\paragraph{Deep Value MPC}
The second formulation (presented by \cite{farshidian2019deep}) extracts both the stage and terminal cost from the learned value function. Hence, DMPC is handle able to sparse and binary rewards.
The intuition behind our modification is that the stage cost represents the value gained between consecutive time steps, and when taken to the limit it can be expressed as as the time integral between two steps of the derivative of the value function with respect to time. Formally, this is referred as the Lie derivative of a function $f(\cdot)$,  $\mathcal{L}_f$. The stage cost then takes the form:
\begin{align}
 g(\boldsymbol{x}_{t}, \boldsymbol{a}_{t}) &:= V_{\theta}(\boldsymbol{x_{t + \Delta T}}) - V_{\theta}(\boldsymbol{x_t}) = \int^{\tau= t + \Delta T}_{\tau = t} \mathcal{L}_f V_{\theta}(\boldsymbol{x}_{t}) \, d\tau = \int^{\tau= t + \Delta T}_{\tau= t} \partial_x V_{\theta}(\boldsymbol{x}_{t})^T \partial_t \boldsymbol{x}_{t} \, d\tau \nonumber\\
&  =  \int^{\tau= t + \Delta T}_{\tau= t}\, \partial_x V_{\theta}(\boldsymbol{x}_{t})^T \hat{f}(\boldsymbol{x}_{t}, \boldsymbol{a}_{t}) \, d\tau \approx \Delta T \, \partial_x V_{\theta}(\boldsymbol{x}_{t})^T \,\hat{f}(\boldsymbol{x}_{t}, \boldsymbol{a}_{t}). 
\end{align}
In \cite{farshidian2019deep}, the authors solve the MPC problem with an algorithm known as SLQ (\cite{sideris2005efficient}), which does not handle constraints systematically and is sensitive to initialization (does not globalize well). SQP on the other hand, efficiently solves both issues. The MPC formulation (Equation~\ref{eq:mpc_cost}) is expressed as:
\begin{equation} \label{eq:deep_mpc_cost}
\pi_{DMPC}(\boldsymbol{x}_0) = \underset{\boldsymbol{a}_{0},\dots, \boldsymbol{a}_{N-1}}{\arg\max} \gamma^{N} V_{\theta}(\boldsymbol{x}_{N}) + \Delta T \sum^{N-1}_{k=0} \gamma^{k} \partial_x V_{\theta}(\boldsymbol{x}_{t})^T \hat{f}(\boldsymbol{x}_{k}, \boldsymbol{a}_{k}).
\end{equation}

\paragraph{Quadratic Program approximation}~\label{sec:qp_approx}
A popular approach to solve non-linear optimization problems is SQP, it solves a sequence of QP approximations of the non-linear problem. We chose this approach to solve the MPC problem due to its ability to handle state-input constraints compared to SLQ. SQP run is time efficient if tailored QP solvers are used it is robust to initial guesses. For our implementation we used the ACADO toolkit toolkit (\cite{Quirynen2014c}) with the QP solver qpOASES (\cite{Ferreau2008}). Note, that ACADO does use the real-time iteration scheme, which only solves one QP per time ste. This further helps to reduce the computational load, for an in depth analysis of the guarantees of this method see \cite{diehl2002real}.

Since the presented value function is differentiable, one could directly find the first and second order derivatives. However, these approximations were found to be numerically unstable, which is due to the fact that the stage cost Hessian uses the third order derivative of the original value function. To mitigate this problem further, a Gauss-Newton approximation of the Hessian has been chosen with a Levenberg-Marquadrdt style damping factor (\cite{marquardt1963algorithm}), which also guarantees that the Hessian is positive semi-definite.

\section{Experiments}\label{sec:experiments}
\subsection{Experimental setup}
\paragraph{Platform} The differential drive UGV used is shown in Figure \ref{fig:teaser}. The on-board visual-inertial SLAM system from Sevensense Robotics\footnote{www.sevensense.ch} provides an estimate of the pose and velocities of this system that are used as feedback to control the robot in real time and to report the results. Training is run on an on-board quadcore Intel i7-6600U CPU at 2.60GHz.
\paragraph{System model} The model used by the MPC controller is a unicycle model with limited velocities and accelerations presented in the following equation:
\begin{align}
	\hat{f}(\boldsymbol{s}_t,\boldsymbol{a}_t) = \left[ \dot{x}, \, \dot{y}, \, \dot{\psi}, \, \dot{v}, \, \dot{\omega} \right]^{T}
	= \left[ v\cos(\psi), \,  v\sin(\psi), \, \omega, \, a, \, \alpha \right]^{T}
\end{align}
where $x$ and $y$ are the position in cartesian coordinates, $\psi$ is the 2D rotation, and $v$ and $\omega$ are the linear and angular velocities in body frame. $a$ and $\alpha$ are the linear and angular accelerations in body frame. Velocities are the input to the UGV, accelerations are only included into the optimization to achieve a smooth behaviour with bounded velocity rates.
\paragraph{Error coordinates} In order to formulate the trajectory tracking objective, we perform a change of coordinates to trajectory error coordinates, where the coordinates are expressed with respect to the reference. Further details can be found in Appendix~\ref{sec:error_coordinates}.

\paragraph{Experiment formulation} 
The objective of all experiments is to accurately track a given reference trajectory. We compare 4 controllers, two learning based , TDMPC, DMPC (see Section \ref{sec:method}) and two classical MPC controllers. The first one is referred to as Naive MPC, which uses the reward as stage cost and terminal cost. Naive MPC serves as the baseline controller for comparision. The second one is referred to as Expert MPC, it was the controller deployed by Sevensense Robotics, where the diagonal weights of the cost function had been hand tuned by their control team.
\subsection{Model mismatch experiment}
\begin{figure}[ht]
\centering
\includegraphics[width=\textwidth]{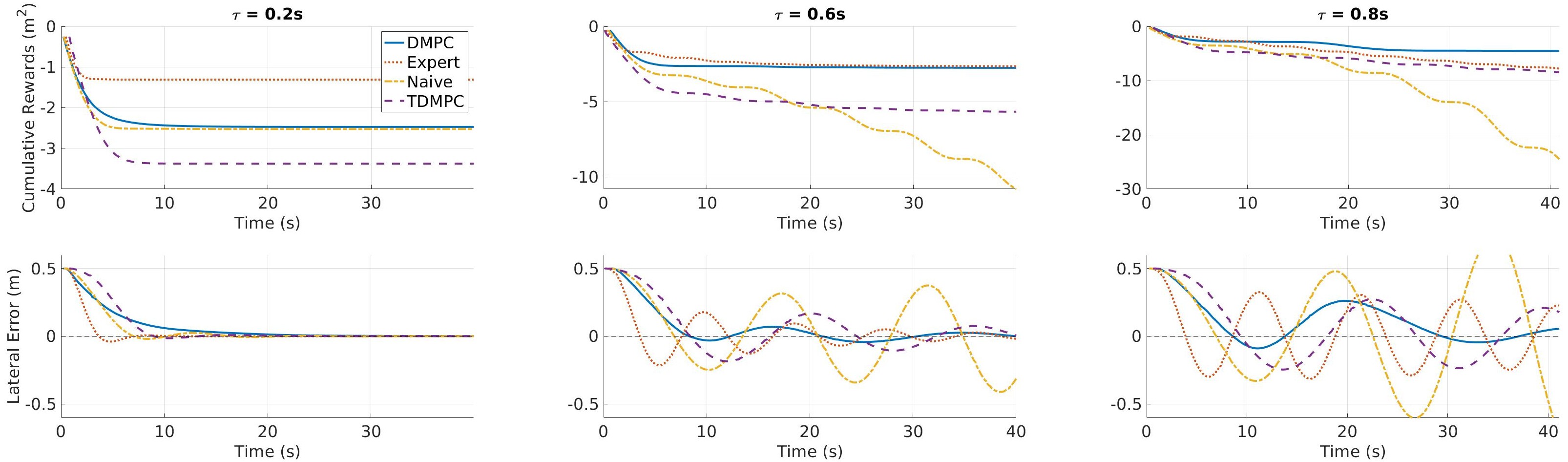}
\caption{Model mismatch experiment. From left to right, the plots show the performance of the methods for increasing time constants \mbox{$\tau = \{0.2, 0.6, 0.8\}s$}. Top row: accumulated rewards over an episode given a dense reward equal to the
tracking error squared. Bottom row: tracking response. DMPC and TDMPC are trained in simulation with $\tau = 0.6s$ but the model still assumes perfect turning dynamics for all methods.}
\label{fig:model_bias}
\vspace{-0.5cm}
\end{figure}

First, we study how the learned value function improves tracking performance in presence of a perturbed system model, without modifying or learning such a model. Consider imperfectly modelled turning dynamics that are represented by a first-order system, where the turning delay is parameterized by a time constant $\tau$.
The learning based methods are trained with $\tau = 0.6s$.
In Figure~\ref{fig:model_bias}, it can be seen that learning based methods perform well even in scenarios far from their training regime; thus learning to be cautious due to the unmodelled dynamics. The Naive and Expert MPC perform well when the model is accurate but quickly deteriorate and even become unstable with $\tau = 0.8s$.

\subsection{Training on a real world UGV}
\begin{figure}[ht]
\centering
\includegraphics[width=\textwidth]{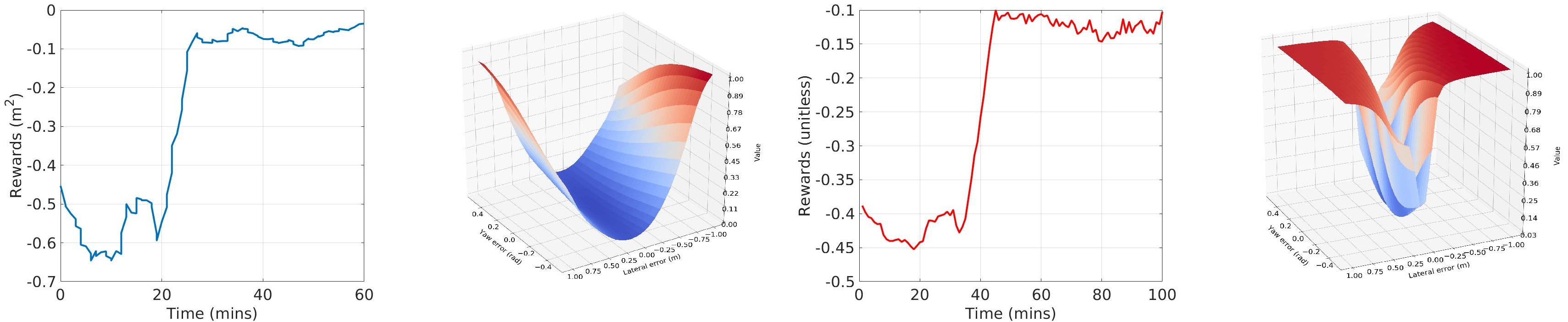}
\caption{Average episode reward over training time with dense rewards (left) along with a 3D representation of the value function and similarly with sparse rewards (right). It can be seen that training with sparse rewards takes longer to converge compared to dense rewards, but still takes less than 45 minutes to converge. Moreover, the shape of the value functions correspond to the rewards given; sparse rewards being much steeper and slimmer compared to dense rewards.}
\label{fig:training_data}
\vspace{-0.5cm}
\end{figure}

TDMPC is capable of learning in the real world on an UGV trained from scratch. This is mostly due to the sample efficiency of the actor i.e TDMPC. It is able to make the learning process converge in within 10000 training iterations, collecting around 3000 samples. In addition, due to the basic knowledge of the model, TDMPC can drive the UGV around the track even if the model is poorly known and the deviations are large at the beginning. The tracking error can be seen in Figure~\ref{fig:training_data} over the training period. Once trained, the learned value function can also be used to run DMPC.

\subsection{Dense reward on real world UGV}
In this experiment, we analyze the performance of all controllers for several paths that cover most typical maneuvers (see Appendix~\ref{sec:experimental_details}). The reward given to the learning based methods is the squared deviation from the path. It can be seen how the Expert MPC behaves best on the real UGV while DMPC consistently outperforms TDMPC. Both learning methods perform consistently better than Naive MPC. It is worth mentioning that DMPC has a damped behavior when approaching the reference in the straight path with no oscillation (see Appendix~\ref{sec:experimental_details}). For the curves and the tight turn scenario, DMPC starts as good as the Expert MPC but eventually accumulates more error.
\begin{figure}[ht]
\centering
\includegraphics[width=\textwidth]{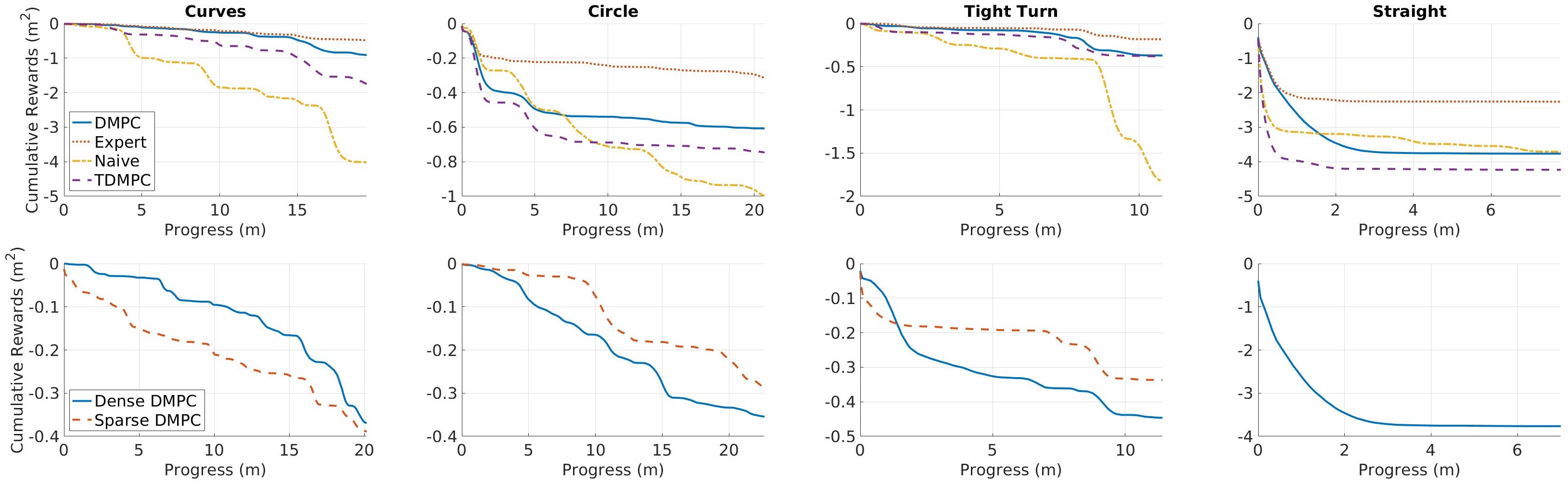}
\caption{Top row: Cumulative rewards over an episode on different trajectories given a dense reward (tracking error squared). Progress means the projected distance travel along the reference path. Bottom row: Cumulative rewards of the DMPC for dense and sparse reward set-up respectively.} \label{fig:benchmarking}
\vspace{-0.5cm}
\end{figure}

\subsection{Sparse rewards on real world UGV} \label{sec:sparse_reward_benchmarking}
Neither Naive, Expert nor TDMPC can handle binary rewards since the cost function has to be differentiable for most efficient solvers. However, DMPC can handle this and we give a binary reward only when close to the reference path. We refer to this as sparse DMPC. This intuitive requirement leads to a successful tracking performance without the need to tune the cost on tracking error. Results can be seen in Figure~\ref{fig:benchmarking}. It can be seen how dense DMPC tracks comparatively worse than sparse DMPC. In this experiment the value function has been trained solely on sparse binary rewards; receiving rewards when the UGV was closer than 10cm to the path (see Appendix \ref{sec:rewards}). It has to be noted that DMPC trained on sparse rewards was not able to complete the straight path experiment starting 0.5m away, as it just stood still. This is probably due to the value function loosing gradient information when very far from the reward. In theory, the reward should be propagated without problems due to the bootstrapping of the target with the value function but in practice the value function flattens out, see Figure \ref{fig:training_data}. 

\section{Related Work}\label{sec:related_work}
Closely related to our work, one can find POLO~(\cite{lowrey2018plan}) and DMPC~(\cite{farshidian2019deep}), which use the learned value function in MPC for the terminal cost, and for the stage and terminal cost, respectively. Our pipeline builds on these works and extends them with practical elements that make it possible to run them on real physical systems. In contrast to this work, \cite{lowrey2018plan} maintains several value functions to encourage exploration. Moreover, \cite{lowrey2018plan} used MPPI (\cite{mppi2017}) as the optimizer. It has the advantage that rewards do not need to be differentiable, as opposed to SQP and SLQ based methods, suffers from the curse of dimensionality. In addition, there are no guarantees that local maxima of the value function will be found. In \cite{farshidian2019deep}, the authors employ a SLQ (\cite{sideris2005efficient}), which is known to scale well with system dimensions, but is very sensitive to the initial guess and it is not trivial to include state constraints.

Another common learning approach in presence of uncertainty is to learn the system's model from data. In this category, Gaussian-Processes (GP) have been successfully used and demonstrated in real and miniature race cars (\cite{juro2019GPs, Hewing2018GP}), but also using deep neural networks (\cite{chua2018nips}). GPs have also been used to model disturbances in order to improve tracking performance (\cite{ostafew2014learning}). Linking value learning and model learning, \cite{gros2019data} pose MPC as a parameterized function approximator, allowing for the entire MPC problem to be subjected to the learning problem. This approach requires for the MPC parameters to be carefully selected for learning. 

There are paradigms where the agent learns from expert demonstration. Notable ones are imitation learning (IM) (\cite{hussein2017imitation}) and inverse reinforcement learning (IRL) (\cite{ng2000algorithms}). It has been shown that combining trajectory optimizion with IM can shorten training time of neural networks (\cite{mordatch2014combining, pmlr-v28-levine13}). A shortcoming of IM and IRL is that the performance of the agent is often bounded by the performance of the expert.

Last but not least, learning based methods are often susceptible to exploring unsafe areas of the state space and it is difficult to guarantee that such methods will not find itself these unsafe areas. Works of \cite{NIPS2017_6692} estimate the region of attraction in order to provide guarantees during the learning process. Applying safe learning to MPC, \cite{koller2018learning} and \cite{rosolia2017learning} construct safe terminal sets based on collected experiences. This could be a natural extension of our work as we currently do not provide any guarantees during training.

\section{Conclusion and Future Work} \label{sec:conclusion}
We presented a sample efficient practical RL scheme capable of training from scratch on a real world physical platform. The algorithm performs value function learning, which is used as the MPC cost function. We use an SQP to solve the corresponding MPC problem. The algorithm is able to learn from given high level objectives including sparse requirements. We demonstrate learned cautious behaviour of the agent in simulation. In practice, it is able to run in real time and matches the performance of expert tuned controllers. The numerical extension presented has shown to be well suited for trajectory tracking. It would prove insightful to explore more challenging scenarios e.g. obstacle avoidance, in future work. Additionally, one could employ a more sophisticated sampling techniques such as prioritized sampling to take advantage of the lesser seen interesting samples.

\acks{The authors thank Farbod Farshidian, Renaud Dubé, Gianluca Cesari and Alex Liniger for their support.}

\bibstyle{bibliography/IEEEtranN}

\newpage
\appendix   

\section{Deep Value MPC Training Algorithm}
\label{sec:training_algorithm}
The Deep Value MPC training process is split into two alternating phases: the roll-out phase and the training phase.
\paragraph{Roll-out phase}
The roll-out phase generates transition samples$\langle\boldsymbol{s}_{t}, \boldsymbol{a}_{t}, r_{t}, \boldsymbol{s}_{t+1}\rangle$, which are stored in the replay buffer $\mathcal{B}$. The samples are generated by taking a control action $\boldsymbol{a}_{t}$ according to $\pi_{TDMPC}(\boldsymbol{s}_{t})$ with the latest learned value function.
\paragraph{Training phase} Using the samples stored in the replay buffer, the target is calculated by using n-step return where the bootstrapping is evaluated using the target network $V_{\theta'}(\boldsymbol{s}_t)$ to stabilize the learning process, specified at equation \ref{eq:n_step_target}. Once the training iteration is complete the value function for MPC and the target network are updated. This process then repeats until convergence.

\begin{algorithm}[ht]
 \SetAlgoLined
  Initialize $\theta, \theta', n, \boldsymbol{s}_{0}$\;\\
  \For{$t = 0 ... T$:}{
   Take action $\boldsymbol{a}_{t} = \pi_{TDMPC}s(_{t})$ with cost function given by equation \ref{eq:terminal_deep_mpc_cost}\\
   Observe new state $\boldsymbol{s}_{t+1} \sim P(\boldsymbol{s}_{t+1}|\boldsymbol{s}_{t}, \boldsymbol{a}_{t})$ and reward $r_{t}$\\
   Store transition tuple in replay buffer $\mathcal{B}$\\
   \If{if mod(t,Z) = 0}{
       Sample mini batch $M$ from replay buffer $\mathcal{B}$\\
       Update critic by minimising loss:\\
       \hspace{0.5cm} $L(\boldsymbol{s}) = \frac{1}{M}\sum_{i}^{M}(R^{n}(\boldsymbol{s}_{i}) - V_{\theta}(\boldsymbol{s}_{i}))^{2} + \beta || \partial_s V_{\theta}(\boldsymbol{s}_{i}) ||_{2}^{2} + \lambda||\theta||_{2}^{2}$\\
       Update target critic:\\
       \hspace{0.5cm}$\theta' \xleftarrow{} \theta'(1-\tau) + \theta\tau$
      }
   }
 \caption{Deep Value MPC Training}
 \end{algorithm}
 
 \section{Reward Engineering}
 \label{sec:rewards}
 \subsection{Dense Rewards}

\begin{equation*}
    r_{dense}(s) = - \left[\left(x_{error}^F\right)^2 + \left(y_{error}^F\right)^2\right]\\
\end{equation*}

\subsection{Sparse Rewards}

\begin{align*}
    r_{sparse}(s) = 
    \begin{cases} 
      -0.5 & \text{if } |x_{error}^F| > 0.1 \text{ or } |\ y_{error}^F| > 0.1 \\
      0 & otherwise  \\
   \end{cases}
\end{align*}{}
 
\section{Error Coordinates}
\label{sec:error_coordinates}
The state of the MDP, $\boldsymbol{s}$, is expressed in terms of frenet or error frame with respect to the reference trajectory and the control point $P$. The high level objective is to track the reference trajectory with the control point.
\begin{figure}[ht]
\centering
\includegraphics[width=8cm]{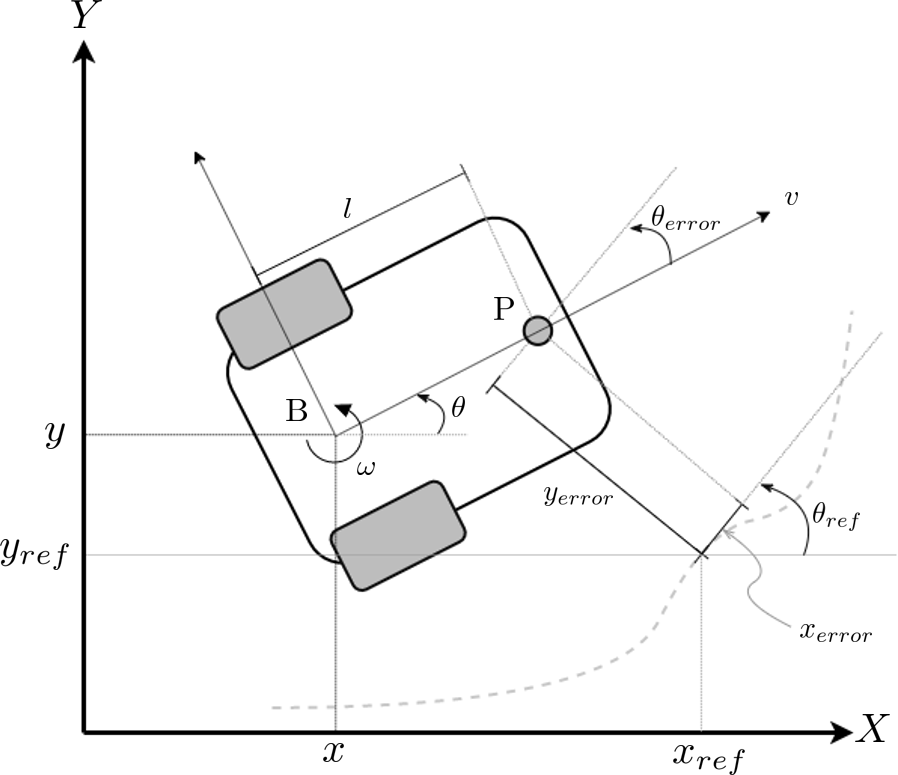}
\caption{Kinematic Differential Drive Model and error coordinates}
\end{figure}

\begin{align*}
 \begin{bmatrix}
 	x_{error}^F\\
 	y_{error}^F\\
  \end{bmatrix} &=
 \begin{bmatrix}
 	x^{M}_{error}\cos(\psi_{ref}) +
    y^{M}_{error}\sin(\psi_{ref})\\
    -x^{M}_{error} \sin(\psi_{ref}) +
    y^{M}_{error}\cos(\psi_{ref})\\
 \end{bmatrix} \,, \,
  \begin{bmatrix}
   x^{M}_{error}\\
   y^{M}_{error}\\
 \end{bmatrix} =
 \begin{bmatrix}
  	x + l\cos(\psi) - x_{ref}\\
  	y + l\sin(\psi) - y_{ref}\\
  \end{bmatrix}
\end{align*}

where $.^{F}$ refers to the frenet frame, $.^{M}$ refers to the map frame. The full frenet states is

\begin{align*}
 \boldsymbol{s}_{error}^F &=
 \begin{bmatrix}
 	x_{error}^F, \ y_{error}^F, \ \psi_{error}^F, \ v, \ \omega
 \end{bmatrix}^{T}
\end{align*}

where $\psi_{error}^F = \psi - \psi_{ref}$. Throughout the paper the subscript are dropped for clarity.

\section{Experimental Details}
In Figure \ref{fig:practical_experiment_details} we show complimentary data on the comparisons of the learning based methods and the classical methods. The first row shows the tracking response of the each controller for each scenario. Second row shows the tracking response but comparing only the dense DMPC with sparse DMPC. It is important to note that we are plotting the performance against the progress along the path rather than time. This way, we are able to compare directly how each controller behave at the same point of the path. Last row shows the reference trajectory for each scenario. The UGV starts at the green dot and finishes at the red dot.
\label{sec:experimental_details}
\subsection{Model mismatch experiment}
The following first order system parameterized by the time constant $\tau$ is used to model the turning delay:
\begin{equation}
\dot{\omega} = \frac{1}{\tau}(\omega_{cmd} - \omega).
\end{equation}

\subsection{Dense reward on a real UGV}
\begin{figure}[ht]
\centering
\includegraphics[width=\textwidth]{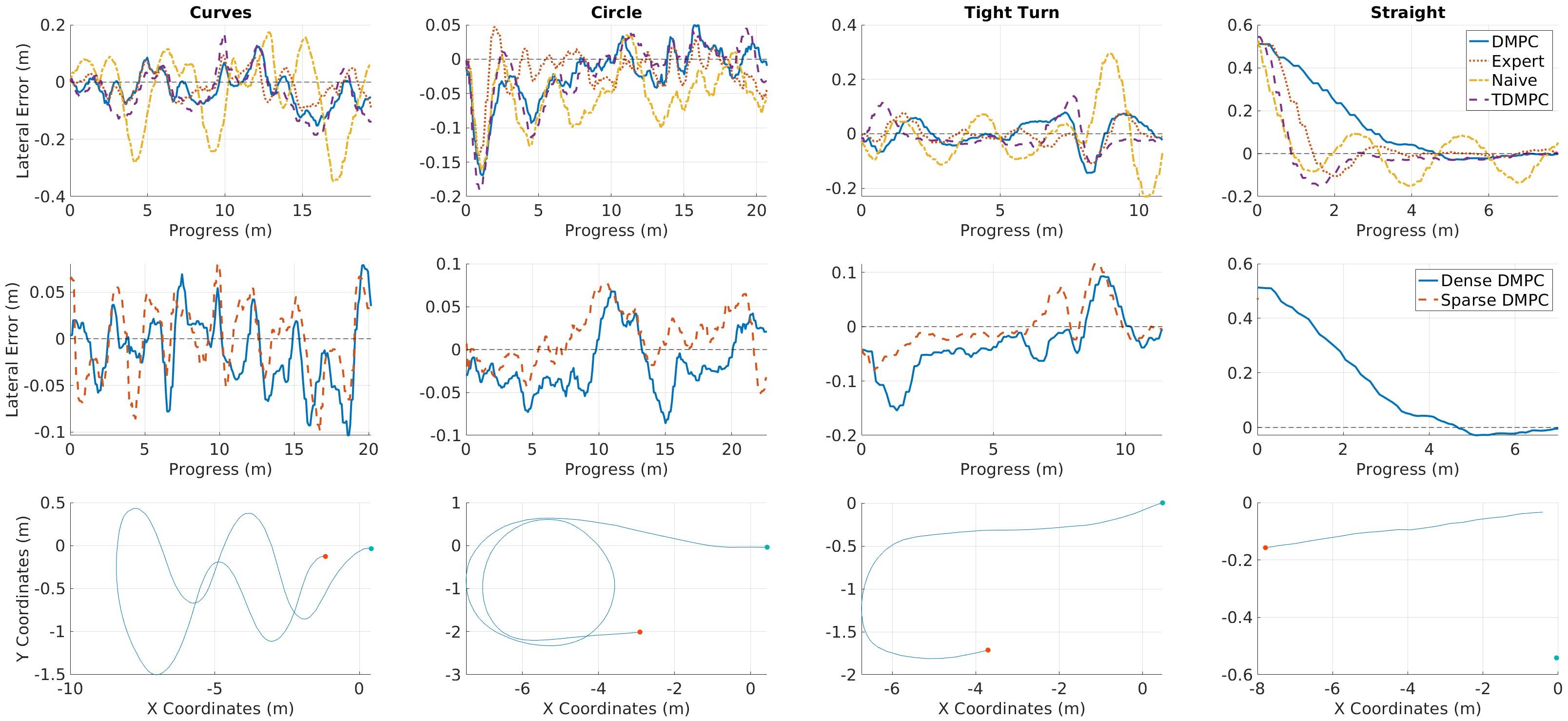} 
\caption{Top row: The lateral tracking error on different trajectories given a dense reward equal to the tracking error squared. Middle row: The lateral tracking error of the DMPC method for dense and sparse reward set-up respectively. Bottom row: The reference trajectories corresponding to each scenario. The UGV starts at the green dot and ends at the red dot.} \label{fig:practical_experiment_details}
\end{figure}

Figure \ref{fig:practical_experiment_details} shows the tracking response on the straight path (top right). Here, the response of the DMPC method is the most damped, approaching the reference in controlled manner with less oscillation. One could argue that this is a more desirable behaviour. This explains why the cumulative rewards of DMPC in Figure \ref{fig:benchmarking} on the straight path seems worse compared to Naive MPC and Expert MPC.
\subsection{Sparse reward on a real UGV}
Reiterating the point from Section \ref{sec:sparse_reward_benchmarking}, the response for sparse reward is missing for the straight scenario because the UGV simply stood still when initialized far from the reference. This is most likely due to the reward not propagating properly resulting in the value function flattening out when far away from reference (see Figure \ref{eq:value_function}). This in turn causes the gradient information to vanish, which makes it difficult for MPC to correctly navigate back to the reference.

\end{document}